\documentclass{article} 
\usepackage{arxiv_style,times}

\usepackage{amsmath,amsfonts,bm}

\def\eqref#1{equation~\ref{#1}}

\def\1{\bm{1}}

\DeclareMathAlphabet{\mathsfit}{\encodingdefault}{\sfdefault}{m}{sl}
\SetMathAlphabet{\mathsfit}{bold}{\encodingdefault}{\sfdefault}{bx}{n}

\usepackage{hyperref}
\usepackage{url}

\providecommand{\coloneqq}{\mathrel{\mathop:}=}

\newcommand{\Reg}{\operatorname{Reg}}

\usepackage{graphicx}
\usepackage{float}
\usepackage{xcolor}

\usepackage{amsmath}
\usepackage{amsthm}
\usepackage{mathtools}
\usepackage{bbm}

\usepackage{algorithm}
\usepackage{algpseudocode}

\makeatletter

\newenvironment{breakablealgorithm}
 {%
 \begin{center}
 \refstepcounter{algorithm}
 \hrule height.8pt depth0pt
 \kern2pt

 \renewcommand{\caption}[2][\relax]{%
  {\raggedright
  \textbf{\ALG@name~\thealgorithm} ##2\par}%

  \ifx\relax##1\relax
  \addcontentsline{loa}{algorithm}{%
   \protect\numberline{\thealgorithm}##2}%
  \else
  \addcontentsline{loa}{algorithm}{%
   \protect\numberline{\thealgorithm}##1}%
  \fi

  \kern2pt
  \hrule
  \kern2pt
 }}
 {%
 \kern2pt
 \hrule
 \relax
 \end{center}
 }

\makeatother

\title{Deep Reinforcement Learning with Buffered Quantile Objectives}

\author{Mohammad Alipour-vaezi\\
Virginia Tech\\
\texttt{alipourvaezi@vt.edu} \\
\And
Sajad Khodadadian\thanks{Corresponding Author} \\
Virginia Tech\\
\texttt{sajadk@vt.edu} \\
}

\begin{document}

\maketitle

\begin{abstract}
Quantile-based reinforcement learning provides an interpretable approach to risk-sensitive decision making by optimizing a prescribed quantile of the cumulative-return distribution. Despite this appeal, learning under a point quantile objective is challenging: quantiles can change abruptly under small perturbations of the return distribution, and exact quantile-sensitive planning requires computationally demanding distributional optimization. Lower-buffered quantiles alleviate the former difficulty by averaging neighboring quantiles immediately below the target level, providing a smoother surrogate while preserving the underlying point-quantile objective. Existing methods based on this principle, however, remain model based and rely on explicit return-law planning, limiting their applicability beyond small tabular problems. We develop \textsc{Deep--BQRL}, a model-free distributional reinforcement-learning framework that extends buffered-quantile learning to neural function approximation. The method learns conditional return quantiles directly from sampled transitions, constructs buffered action scores from the relevant region of the learned quantile function, and uses ensemble disagreement to guide exploration. An augmented input representation allows the learned policy to respond to trajectory information without explicitly reproducing the quantile-state recursion required by exact planning. Experiments on an asset-selling optimal-stopping problem and slippery FrozenLake compare \textsc{Deep--BQRL} with model-based \textsc{UCB--BQRL} and tabular PPO and TRPO implementations. In asset selling, \textsc{Deep--BQRL} attains smaller mean cumulative point-quantile policy gaps than PPO and TRPO at the reported target levels, while \textsc{UCB--BQRL} retains the
smallest gaps. The learned stopping decisions also vary with the target quantile, providing an interpretable illustration of the method's risk-sensitive behavior.
\end{abstract}

\section{Introduction}
\label{sec:introduction}

Reinforcement learning (RL) is commonly formulated to maximize expected cumulative reward \citep{sutton1998reinforcement}. In many applications, however, the expectation alone does not adequately characterize performance: two policies with similar mean returns may have substantially different probabilities of poor outcomes \citep{li2022quantile}. This motivates \emph{risk-sensitive} RL, where decisions depend on the distribution of cumulative return rather than only its expectation \citep{howard1972risk,garcia2015comprehensive,gottesman2019guidelines}. Quantile objectives provide a particularly interpretable criterion by directly targeting a specified region of the return distribution \citep{filar1995percentile,delage2010percentile,li2022quantile}. In RL, maximizing the $\tau$-quantile of cumulative reward seeks a policy that makes the return at the $\tau$-th percentile of its cumulative-return distribution as large as possible.

Optimizing return quantiles nevertheless poses important statistical and computational challenges. Point quantiles can be sensitive to small perturbations of the underlying return distribution, making them difficult to estimate reliably when the environment is learned from finite data \citep{alipour-vaezi2026optimistic}. To mitigate this instability, \citet{alipour2026risk} introduced the lower-buffered quantile within the \textsc{UCB--BQRL} framework, which averages quantiles over a neighborhood immediately below the target level. This criterion retains the distributional interpretation of a quantile while smoothing its dependence on the return law. It is closely related to Range Value-at-Risk and interpolates between a point quantile and a lower-tail average as the buffer width varies \citep{fissler2021elicitability,embrechts2018quantile}.

The remaining difficulty is scalability. Exact quantile planning is fundamentally different from ordinary Bellman optimization because the optimal decision may depend on the realized trajectory through a dynamically evolving quantile level \citep{li2022quantile}. The model-based \textsc{UCB--BQRL} framework provides a principled tabular solution for learning under buffered-quantile objectives, but its planner requires explicit manipulation of return distributions and optimization over transition confidence sets. Moreover, exact evaluation of point and buffered quantiles is computationally hard in general \citep{alipour2026risk}. These operations become increasingly costly as the state and action spaces grow and motivate a model-free neural approximation that avoids explicit return-law planning during training.

We develop \textsc{Deep--BQRL}, a distributional deep RL framework for risk-sensitive control under lower-buffered quantile objectives. Rather than estimating a transition model and repeatedly solving a quantile-planning problem, \textsc{Deep--BQRL} learns a fixed collection of conditional return quantiles directly using distributional critics \citep{bellemare2017distributional,dabney2018distributional}. The buffered action score is then obtained by aggregating the learned quantiles over $[\tau-\beta,\tau]$, allowing policy improvement to depend on a local region of the return distribution rather than on a single estimated quantile. To account for the history dependence of quantile-sensitive decisions \citep{li2022quantile}, the policy and critic operate on an augmented state representation that incorporates trajectory information. We further maintain an ensemble of distributional critics and use disagreement in their buffered action scores to encourage exploration directly in return-value space.

Our contributions are threefold. First, we develop \textsc{Deep--BQRL}, a model-free deep RL method for quantile-sensitive decision making that avoids learning an explicit transition model and repeatedly solving an exact quantile-planning problem. Second, we use learned return quantiles to construct a lower-buffered criterion for action selection, providing a practical way to use the smoother objective with neural function approximation. Third, we use disagreement among multiple learned critics to encourage exploration in parts of the return distribution that remain uncertain. The main experiment uses an asset-selling (optimal-stopping) problem \citep{seierstad1992reservation,rosenfield1983optimal} as well as slippery FrozenLake to compare learning performance and examine how the target quantile affects the learned stopping decisions.

\section{Preliminaries}
\label{sec:preliminaries}

\subsection{Finite-Horizon Reinforcement Learning}

We consider a finite-horizon Markov decision process $\mathcal M=(\mathcal S,\mathcal A,H,P,r)$, where $\mathcal S$ and $\mathcal A$ are the state and action spaces, $H$ is the horizon, $P=\{P_h\}_{h=0}^{H-1}$ is the collection of stage-dependent transition kernels, and $r=\{r_h\}_{h=0}^{H-1}$ is the collection of bounded one-step reward functions. At stage $h$, the agent observes state $S_h$, selects action $A_h$, transitions according to $S_{h+1}\sim P_h(\cdot\mid S_h,A_h)$, and receives $R_h=r_h(S_h,A_h,S_{h+1})$. Thus, $r_h(s,a,s')$ is deterministic conditional on a realized transition $(s,a,s')$, while $R_h$ may be random through the random next state $S_{h+1}$. We assume a fixed initial-state distribution and suppress it from the notation. A policy $\pi=\{\pi_h\}_{h=0}^{H-1}$ may in general depend on the observed trajectory. Its total episodic return is $R^\pi\coloneqq\sum_{h=0}^{H-1}R_h$.

\subsection{Quantile Objective and Lower-Buffered Quantile}

For a bounded real-valued random variable $W$ and $q\in(0,1]$, its $q$-quantile is $Q_q(W)\coloneqq\inf\{w\in\mathbb R:\Pr(W\le w)\ge q\}$. For completeness, we use $Q_0(W)\coloneqq\inf\operatorname{supp}(W)$. For a target level $\tau\in(0,1)$, the risk-sensitive objective of interest is $\max_\pi Q_\tau(R^\pi)$. Thus, the goal is to find a policy whose cumulative return is as large as possible at the prescribed quantile level. Smaller values of $\tau$ emphasize lower-tail outcomes, whereas larger values place greater emphasis on upper-tail realizations \citep{li2022quantile}. Direct optimization of this objective is challenging because point quantiles can change abruptly under small perturbations of the return distribution. To obtain a smoother criterion, following \citet{alipour2026risk}, for a buffer width $\beta\in(0,\tau]$ we define the lower-buffered quantile $Q_\tau^\beta(W)\coloneqq\frac{1}{\beta}\int_{\tau-\beta}^{\tau} Q_u(W)\,du$. Rather than relying on a single point of the quantile function, $Q_\tau^\beta(W)$ averages quantiles over the interval immediately below $\tau$. As $\beta$ decreases, the buffered criterion approaches the point quantile, while larger values of $\beta$ incorporate more information from neighboring lower quantiles \citep{alipour2026risk}.

\section{\textsc{Deep--BQRL}}
\label{sec:deep-bqrl}
We now introduce \textsc{Deep--BQRL}, a model-free distributional learning method for finite action spaces. A key distinction between expected-return and quantile-sensitive control is that the optimal policy of a quantile-sensitive decision is history-dependent, i.e., it depends on the entire trajectory and not only on the current state. Exact QMDP planning handles this dependence by updating an auxiliary quantile level after each observed transition \citep{li2022quantile}. Exact lower-buffered planning similarly relies on the notion of return distributions, which captures history dependence of the sampling policy \citep{alipour2026risk}. Both operations become computationally demanding beyond small tabular problems. Therefore, \textsc{Deep--BQRL} uses three approximations. First, Approximation I compresses trajectory information into the current stage, state, and cumulative reward. Second, Approximation II represents each conditional remaining-return distribution by a finite collection of learned quantiles. Third, Approximation III replaces exact quantile planning with greedy policy improvement by its approximation, denoted as buffered action scores. 

\paragraph{Approximation I: Augmented Trajectory Representation}
The first approximation replaces the trajectory information with a fixed-dimensional trajectory summary. At stage $h$, let $C_h\coloneqq\sum_{k=0}^{h-1}R_k$, denote the cumulative reward collected before action $A_h$ is selected. We define the augmented input $X_h\coloneqq(h,S_h,C_h)$, $h=0,\ldots,H$, and write $x=(h,s,c)$ for a realization of $X_h$. The augmented input is an implemented trajectory summary.  For $x=(h,s,c)$ and $a\in\mathcal A$, let $Z^\pi(x,a)$ denote a random variable with the same law as the remaining reward $\sum_{k=h}^{H-1}R_k$ when action $a$ is taken at augmented input $x$, and policy $\pi$ is followed thereafter. At $x$, the previously collected reward $c$ is already known, so the corresponding total return has the same law as $c+Z^\pi(x,a)$. The critic estimates the distribution of the remaining return $Z^\pi(x,a)$, taking $c$ as part of the input. This is because a history-dependent policy may condition its future actions on the reward accumulated so far.

\paragraph{Approximation II: Quantile Representation of Remaining Returns}
The second approximation estimates the law of $Z^\pi(x,a)$, by a finite collection of learned quantiles. \textsc{Deep--BQRL} maintains $M\ge2$ separately initialized critics. For the fixed levels $\eta_j\coloneqq(j-\frac12)/K$, $j=1,\ldots,K$, critic $m$ estimates $z_{\theta_m,j}(x,a) \approx Q_{\eta_j}\!\left(Z^\pi(x,a)\right)$. Quantile-regression distributional RL provides a practical mechanism
for learning these quantiles from sampled transitions
\citep{bellemare2017distributional,dabney2018distributional}. Because the critic can occasionally predict a larger return for a lower quantile level than for a higher one, we sort the $K$ predicted returns for each state--action pair. Let $\widetilde z_{\theta_m,1}(x,a) \le \cdots \le \widetilde z_{\theta_m,K}(x,a)$ denote these sorted outputs. The sorted values are used only in subsequent decision-making and learning targets. When training the critic, the original output $z_{\theta_m,j}(x,a)$ remains associated with its fixed quantile level $\eta_j$.

\paragraph{Approximation III: Buffered Policy Improvement}
For a given augmented input $x$ and action $a$, the lower-buffered value of the remaining return is $Q_\tau^\beta\!\left(Z^\pi(x,a)\right) = \frac{1}{\beta}
\int_{\tau-\beta}^{\tau} Q_u\!\left(Z^\pi(x,a)\right)\,du$. This quantity averages the return quantiles between $\tau-\beta$ and $\tau$. Ideally, actions would be compared using $Q_\tau^\beta\!\left(Z^\pi(x,a)\right)$. In \textsc{Deep--BQRL}, however, the full quantile function $Q_u(Z^\pi(x,a))$ is not available. Approximation~II provides only $K$ predicted quantiles. Each level $\eta_j$ corresponds to the interval $[(j-1)/K,j/K]$. We therefore use the predicted quantile at $\eta_j$ to represent the portion of this interval that lies inside $[\tau-\beta,\tau]$. The length of this overlap is
\begin{equation}
\label{eq:buffer-weight}
w_j^{\tau,\beta}
\coloneqq
\left[
\min\left\{\tau,\frac{j}{K}\right\}
-
\max\left\{\tau-\beta,\frac{j-1}{K}\right\}
\right]_+,
\qquad j=1,\ldots,K,
\end{equation}
where $[v]_+\coloneqq\max\{v,0\}$. Using the sorted predicted returns from critic $m$, we approximate $Q_\tau^\beta\!\left(Z^\pi(x,a)\right)$ by $\widehat B_m(x,a) \coloneqq \frac{1}{\beta} \sum_{j=1}^{K}
w_j^{\tau,\beta} \widetilde z_{\theta_m,j}(x,a)$. Thus, $\widehat B_m(x,a)$ is the buffered score assigned to action $a$ by critic $m$. Because \textsc{Deep--BQRL} maintains $M$ critics, we average their
scores: $\mu_B(x,a) \coloneqq \frac{1}{M} \sum_{m=1}^{M} \widehat B_m(x,a)$. After training, the policy selects the action with the largest ensemble-mean buffered score:
\begin{equation}
\label{eq:deep-bqrl-output-policy}
\widehat\pi(x)
\in
\arg\max_{a\in\mathcal A}
\mu_B(x,a).
\end{equation}

\subsection{Critic Learning and Exploration}
\label{subsec:critic-learning}

The agent stores observed transitions in a replay buffer $\mathcal D$. A replay sample is denoted by $(x,a,r,x',d)$, where $x$ and $x'$ are consecutive augmented inputs, $a$ is the selected action, $r$ is the observed reward, and $d\in\{0,1\}$. We set $d=1$ when the environment terminates or when the transition reaches the finite-horizon boundary; otherwise, $d=0$. To stabilize training, for each critic $z_{\theta_m}$ we maintain a slowly updated copy $z_{\bar\theta_m}$, called the target critic. The target critic is used to evaluate the next augmented input when constructing the training targets.

For each critic $m$, we first evaluate its target critic $z_{\bar\theta_m}$ at the next augmented input $x'$ and use these predictions to select the next action. After sorting the target-critic outputs, define $\overline B_m(x',a)\coloneqq \frac{1}{\beta} \sum_{j=1}^{K} w_j^{\tau,\beta}\widetilde z_{\bar\theta_m,j}(x',a)$. For a nonterminal next input $x'$, critic $m$ selects
\begin{equation}
\label{eq:target-next-action}
a_m^+(x')
\in
\arg\max_{a\in\mathcal A}
\overline B_m(x',a).
\end{equation}
The corresponding target return values are $y_{m,i} \coloneqq r + (1-d) \widetilde z_{\bar\theta_m,i} \!\left(x',a_m^+(x')\right) $, $i=1,\ldots,K$. When $d=1$, all target values reduce to $y_{m,i}=r$, and the continuation action need not be evaluated.

Following the Huber loss of \citet{huber1992robust} and its quantile-regression formulation in distributional RL \citep{dabney2018distributional}, we train each critic using the quantile Huber loss. For a scalar residual $\delta$ and threshold $\kappa>0$, define the Huber loss
\begin{equation*}
\label{eq:huber-penalty}
h_\kappa(\delta)
\coloneqq
\begin{cases}
\frac12\delta^2,
& |\delta|\le\kappa,\\[1mm]
\kappa\left(|\delta|-\frac12\kappa\right),
& |\delta|>\kappa.
\end{cases}
\end{equation*}
The corresponding quantile Huber loss is $\rho_{\eta_j}^{\kappa}(\delta) \coloneqq \left|\eta_j- \mathbbm 1\{\delta<0\}\right|\frac{h_\kappa(\delta)}{\kappa}$. The Huber loss is quadratic for small residuals and linear for large residuals, while the weight $\left|\eta_j-\mathbbm 1\{\delta<0\}\right|$ in quantile Huber loss directs the prediction toward the $\eta_j$-quantile of the target return distribution. 

For a minibatch $\mathcal B\subset\mathcal D$, critic $m$ minimizes
\begin{equation}
\label{eq:deep-bqrl-member-loss}
  \mathcal J_m(\theta_m)
  \coloneqq
  \frac{1}{|\mathcal B|K^2}
  \sum_{(x,a,r,x',d)\in\mathcal B}
  \sum_{i=1}^{K}
  \sum_{j=1}^{K}
  \rho_{\eta_j}^{\kappa}
  \left(
    y_{m,i}
    -
    z_{\theta_m,j}(x,a)
  \right).
\end{equation}
Each ensemble member is updated by minimizing its corresponding quantile-regression loss $\mathcal J_m(\theta_m)$. This is the quantile-regression distributional Bellman loss \citep{dabney2018distributional}, applied to the augmented trajectory representation. Minimizing this loss updates the critic parameters so that $z_{\theta_m,j}(x,a)$ approximates the $\eta_j$-quantile of the conditional remaining-return distribution. The parameters $(\tau,\beta)$ do not appear explicitly in the quantile Huber loss; they affect training through the buffered scores used to select actions and construct the target values.

After each gradient update, the target parameters are updated using a soft target-network update \citep{lillicrap2015continuous},
\begin{equation}
\label{eq:target-update}
  \bar\theta_m
  \leftarrow
  (1-\zeta)\bar\theta_m+\zeta\theta_m,
  \qquad m=1,\ldots,M,
\end{equation}
where $\zeta\in(0,1]$ controls the update rate.

During data collection, \textsc{Deep--BQRL} encourages exploration by adding a bonus to actions for which the critics disagree more strongly about the buffered score. Define $\sigma_B(x,a) \coloneqq [
\frac{1}{M-1} \sum_{m=1}^{M} (\widehat B_m(x,a)-\mu_B(x,a))^2 ]^{1/2}$. During episode $t$, the behavior policy selects $A_h \in \arg\max_{a\in\mathcal A} \left\{\mu_B(X_h,a)+ \lambda_t\sigma_B(X_h,a)
\right\}$, where $\lambda_t \coloneqq \frac{\lambda_0} {1+t/\lambda_{\mathrm{decay}}}$ is the exploration coefficient. 

Algorithm~\ref{alg:deep-bqrl} combines the augmented trajectory representation, the quantile representation of remaining returns, and buffered policy improvement. Quantile regression trains the critics, while ensemble disagreement guides data collection.

\begin{breakablealgorithm}
\caption{\textsc{Deep--BQRL} for discrete-action buffered-quantile control}
\label{alg:deep-bqrl}
\begin{algorithmic}[1]

\State \textbf{Input:} target level $\tau\in(0,1)$; buffer
$\beta\in(0,\tau]$; number of quantiles $K$; number of critics
$M\ge2$; Huber threshold $\kappa$; target-update rate $\zeta$;
exploration parameters $\lambda_0$ and $\lambda_{\mathrm{decay}}$.

\State Initialize replay buffer $\mathcal D$, critics
$\{z_{\theta_m}\}_{m=1}^{M}$, and target critics
$\{z_{\bar\theta_m}\}_{m=1}^{M}$ with
$\bar\theta_m\gets\theta_m$.

\State Compute the weights
$\{w_j^{\tau,\beta}\}_{j=1}^{K}$ using
\eqref{eq:buffer-weight}.

\For{each training episode $t$}

  \State Observe the initial state $s$ and set $x\gets(0,s,0)$.

  \For{$h=0,\ldots,H-1$}

    \State Compute $\widehat B_m(x,a')$ $\forall m\!\in\!\{\!1,\!\ldots\!,\!M\!\}$ and $a'\!\in\!\mathcal A$. Then compute
    $\mu_B(x,a')$ and $\sigma_B(x,a')$.

    \State Select
    $a\in\arg\max_{a'\in\mathcal A}
    \left\{\mu_B(x,a')+\lambda_t\sigma_B(x,a')\right\}$.

    \State Execute $a$ and observe reward $r$ and next state $s'$.

    \State Set $d=1$ if the episode terminates or $h=H-1$;
    otherwise set $d=0$.

    \State If $x=(h,s,c)$, set
    $x'\gets(h+1,s',c+r)$.

    \State Store $(x,a,r,x',d)$ in $\mathcal D$.

    \If{$\mathcal D$ contains enough samples for a minibatch}

      \State Sample $\mathcal B\subset\mathcal D$.

      \For{$m=1,\ldots,M$}

        \State For each $(x,a,r,x',d)\in\mathcal B$, evaluate and
        sort the target-critic outputs at $x'$.

        \If{$d=0$}
          \State Compute $\overline B_m(x',a')$ $\forall a'\in\mathcal A$ and select $a_m^+(x')$ using
          \eqref{eq:target-next-action}.
          \State Construct $\{y_{m,i}\}_{i=1}^{K}$ using the target
          values defined above.
        \Else
          \State Set $y_{m,i}\gets r$ for $i=1,\ldots,K$.
        \EndIf

        \State Update $\theta_m$ by minimizing
        $\mathcal J_m(\theta_m)$ in
        \eqref{eq:deep-bqrl-member-loss}.

        \State Update $\bar\theta_m$ using
        \eqref{eq:target-update}.

      \EndFor

    \EndIf

    \If{$d=1$}
      \State \textbf{break}
    \Else
      \State $x\gets x'$
    \EndIf

  \EndFor
\EndFor

\State \textbf{Output:} $\widehat\pi$ defined in
\eqref{eq:deep-bqrl-output-policy}.

\end{algorithmic}
\end{breakablealgorithm}

\section{Results}
\label{sec:results}

We evaluate \textsc{Deep--BQRL} on the finite-horizon asset-selling problem \citep{seierstad1992reservation,rosenfield1983optimal} studied by \citet{alipour2026risk}. At each decision period, a seller observes an offer and chooses whether to accept it or wait for another. Accepting yields a known reward, while waiting offers the possibility of a higher reward but also the risk of receiving less. This makes the problem useful for evaluating both learning performance and how different target quantiles affect decisions. We also evaluate \textsc{Deep--BQRL} on slippery FrozenLake, a stochastic navigation problem with sparse terminal rewards, to test the method under a different transition structure.

We compare \textsc{Deep--BQRL} with the model-based \textsc{UCB--BQRL} method \citep{alipour2026risk} and the PPO and TRPO implementations used in our experiments \citep{schulman2017proximal,schulman2015trust}. The state space in both environments are small enough for the model-based method to remain computationally feasible, providing a reference for evaluating the model-free approach. For both environments, hyperparameters are selected independently of the held-out evaluation runs. Following \citet{alipour2026risk}, structural choices are fixed, and one dominant hyperparameter per method is varied over a $16$-point logarithmic grid. We use successive halving with reduction factor $\eta=2$ and increasing training budgets. Validation uses seeds $1000$, $1001$, and $1002$; configurations are ranked primarily by cumulative $\tau$-quantile policy performance, with late-training episodic return as a tie-breaker. Final results use the disjoint seeds $42$, $10042$, and $20042$.


\subsection{Asset-Selling}
\label{subsec:asset-selling}

The environment has $25$ offer states, $s\in\{0,\ldots,24\}$, and an absorbing post-sale state. Each episode starts at $S_0=5$ and allows at most $H=10$ decisions. At each state, the action set is $\mathcal A=\{\mathrm{Stop},\mathrm{Continue}\}$. Choosing $\mathrm{Stop}$ accepts the current offer, yields reward $s/24$, and ends the episode; division by $24$ normalizes the sale reward to $[0,1]$. Choosing $\mathrm{Continue}$ rejects the offer, yields zero immediate reward, and draws a new offer independently and uniformly from $\{0,\ldots,24\}$. If no offer is accepted within the horizon, the episode ends with zero return. Thus, the total episodic return is the normalized accepted offer, or zero if no sale occurs. The decision trades off accepting the current offer against waiting for an uncertain future offer. A lower target quantile emphasizes protection against low sale rewards, whereas a higher target quantile emphasizes more favorable outcomes. The learned policy determines how this preference affects stopping decisions across offers and decision periods. These experiments use $2{,}000$ training episodes per run and three independent seeds. We report learning curves for $\tau\in\{0.1,0.9\}$ and selected final policies for $\tau\in\{0.1,0.5,0.9\}$. Curves show the mean across the three runs, with shaded pointwise $95\%$ Student-$t$ intervals. Given only three runs, these intervals summarize variability rather than provide evidence of statistically significant differences.

We report two cumulative performance measures. Let $G_t$ denote the realized return in training episode $t$. The empirical cumulative expected-return regret is $\widehat{\Reg}_{E}(k)\coloneqq\sum_{t=1}^{k}\left(\max_{\pi}\mathbb E[R^\pi]-G_t\right)$, which compares rewards collected by the behavior policy, including exploration, with the optimal expected episodic return under the true environment. To evaluate the learned policy without the exploration bonus, let $\bar\pi_t$ denote the most recently evaluated greedy policy available at episode $t$. We report the cumulative $\tau$-quantile policy gap $\widehat{\mathcal G}_{\tau}(k)\coloneqq\sum_{t=1}^{k}\left[Q_\tau^\star-Q_\tau(R^{\bar\pi_t})\right]_+$, where $Q_\tau^\star$ is the common numerical point-quantile reference computed from the true environment model. Policy evaluation is periodic, and the latest evaluated value is retained between evaluations. The two measures capture different aspects of learning: expected-return regret reflects rewards collected during training, while the quantile policy gap reflects the quality of the evaluated greedy policy at the prescribed quantile level, so their magnitudes should not be compared directly. Although \textsc{Deep--BQRL} selects actions using the lower-buffered criterion $Q_\tau^\beta$, the policy-gap measure evaluates the target point quantile $Q_\tau$.

\begin{figure}[h]
\centering

\includegraphics{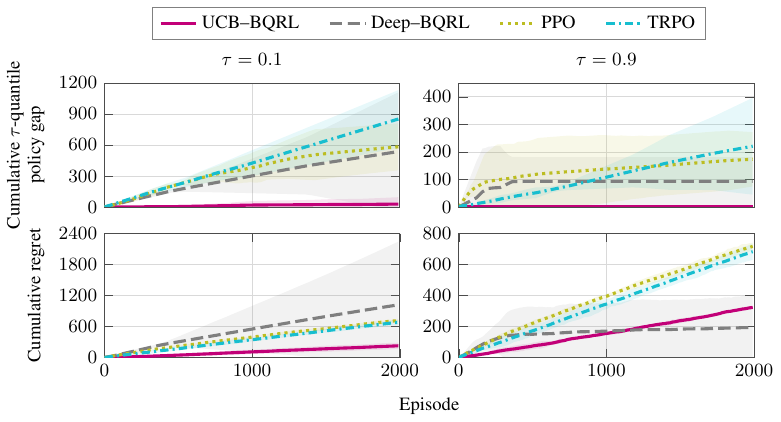}

\caption{
Learning performance on the asset-selling environment for
$\tau\in\{0.1,0.9\}$. The top row reports the cumulative
$\tau$-quantile policy gap, while the bottom row reports cumulative
regret under the expected-return criterion. Curves show the mean over
three independent seeds; shaded regions denote pointwise $95\%$
Student-$t$ intervals. Lower values indicate better performance.
}
\label{fig:asset-selling-performance}
\end{figure}

Figure~\ref{fig:asset-selling-performance} compares the cumulative measures for $\tau=0.1$ and $\tau=0.9$. At episode $2000$, the mean cumulative point-quantile gap of \textsc{Deep--BQRL} is $538.33$ for $\tau=0.1$, compared with $583.63$ for PPO and $855.22$ for TRPO; for $\tau=0.9$, the corresponding values are $92.78$, $173.51$, and $221.36$. Thus, \textsc{Deep--BQRL} has smaller gaps than both policy-gradient baselines at both target levels, while model-based \textsc{UCB--BQRL} remains substantially stronger, with gaps of $28.14$ and $0.79$. Expected-return regret shows a different pattern. For $\tau=0.1$, \textsc{Deep--BQRL} reaches $1025.44$, compared with $229.00$ for \textsc{UCB--BQRL}, $719.58$ for PPO, and $685.96$ for TRPO. For $\tau=0.9$, \textsc{Deep--BQRL} has the smallest regret, $193.68$, compared with $324.65$, $719.58$, and $685.96$, respectively. Thus, a smaller point-quantile gap does not necessarily imply greater reward during training: \textsc{Deep--BQRL} improves on PPO and TRPO under the point-quantile measure, while its expected-return performance depends on the target level.

\begin{figure}[t]
\centering

\includegraphics{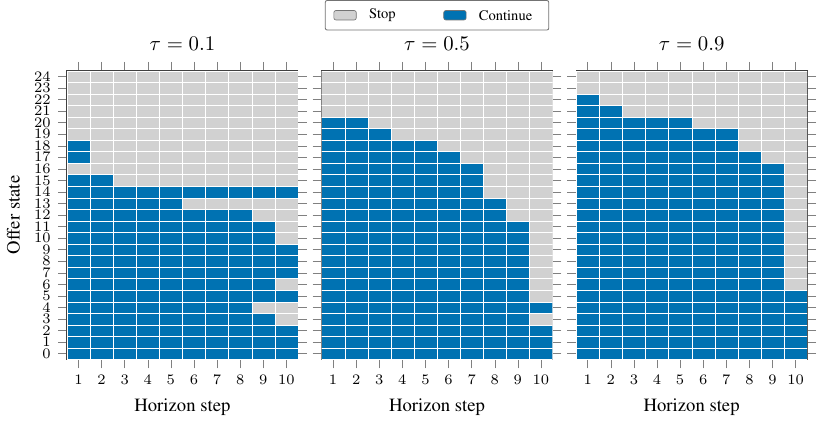}

\caption{Selected final \textsc{Deep--BQRL} policies in the asset-selling
environment for $\tau\in\{0.1,0.5,0.9 \}$, evaluated at $C_h=0$.
Blue cells indicate $\mathrm{Continue}$ and gray cells indicate
$\mathrm{Stop}$. The panels show individual runs with seeds $10042$,
$42$, and $20042$, respectively, rather than policies averaged across
runs. The absorbing post-sale state is omitted.
}
\label{fig:asset-selling-deep-policy}
\end{figure}

Figure~\ref{fig:asset-selling-deep-policy} shows selected final \textsc{Deep--BQRL} policies, where blue denotes $\mathrm{Continue}$ and gray denotes $\mathrm{Stop}$. Each panel shows one run, with seeds $10042$, $42$, and $20042$ for $\tau=0.1$, $0.5$, and $0.9$, respectively. Overall, smaller target quantiles lead to accepting lower offers, while larger target quantiles favor waiting for higher offers. For example, in the second decision period, the $\tau=0.1$, $0.5$, and $0.9$ policies accept offers of at least $16$, $21$, and $22$, respectively, illustrating the shift from protecting against low returns to pursuing more favorable outcomes. The willingness to wait generally decreases toward the horizon, although the learned boundaries are not perfectly regular. Some positive final-period offers are still rejected despite no remaining sale opportunity (e.g., at $h=10$, offers $1$--$9$, $1$--$4$, and $1$--$5$ for $\tau=0.1$, $0.5$, and $0.9$, respectively). Thus, the figure shows learned approximations rather than exact optimal stopping rules.

\subsection{Slippery FrozenLake}
\label{subsec:frozenlake}

The standard $4\times4$ slippery FrozenLake environment has $16$
states and actions
$\mathcal A=\{\textsc{Left},\textsc{Down},\textsc{Right},\textsc{Up}\}$.
Each episode begins at a fixed start state and lasts at most
$H=100$ steps. Frozen cells are safe, entering a hole ends the
episode with zero reward, and reaching the goal ends it with reward
one; all other rewards are zero. Movement is stochastic under the
slippery dynamics. Thus, $R^\pi\in\{0,1\}$, and the expected episodic
return equals the probability of reaching the goal before entering
a hole or reaching the horizon. We report $5{,}000$ training episodes per run for
$\tau\in\{0.1,0.5,0.9\}$. Curves show means across the three held-out
seeds, with shaded pointwise $95\%$ Student-$t$ intervals interpreted
as descriptive summaries of variability. We use the two cumulative
measures defined in Section~\ref{subsec:asset-selling} and additionally
report a $50$-episode moving average of $G_t$, which summarizes the
observed frequency of reaching the goal during training.

\begin{figure}[ht]
\centering

\includegraphics{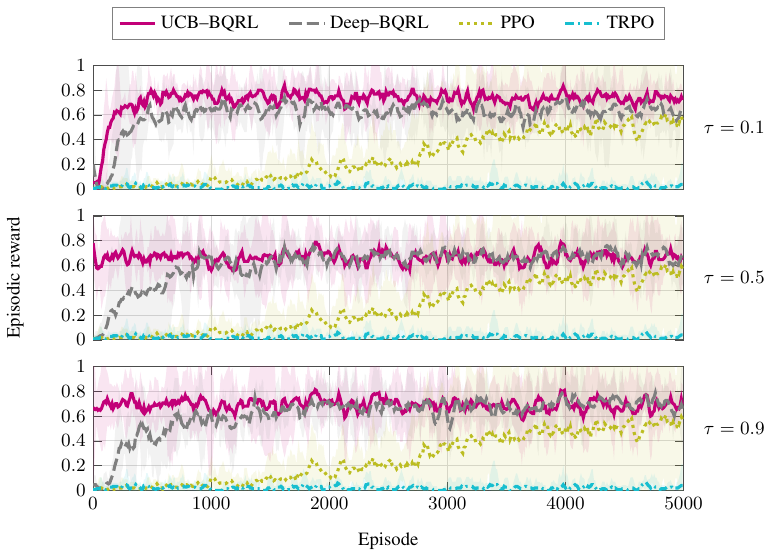}

\caption{
Episodic reward on the slippery FrozenLake environment for $\tau\in\{0.1,0.5,0.9\}$. Curves report the 50-episode moving-average reward averaged over three independent seeds; shaded regions denote pointwise 95\% Student-$t$ confidence intervals.
}
\label{fig:frozenlake-slippery-reward}
\end{figure}

Figure~\ref{fig:frozenlake-slippery-reward} shows faster initial
learning for \textsc{UCB--BQRL}, while \textsc{Deep--BQRL} improves
more gradually. At episode $5000$, the mean $50$-episode
moving-average rewards of \textsc{Deep--BQRL} are $0.627$, $0.687$,
and $0.673$ for $\tau=0.1$, $0.5$, and $0.9$, respectively,
compared with $0.713$, $0.640$, and $0.680$ for
\textsc{UCB--BQRL}. PPO reaches approximately $0.51$--$0.53$,
whereas TRPO remains near zero. Thus, \textsc{Deep--BQRL} achieves
episodic reward close to the model-based reference, with a
numerically higher terminal mean at $\tau=0.5$.

\begin{figure}[ht]
\centering

\includegraphics{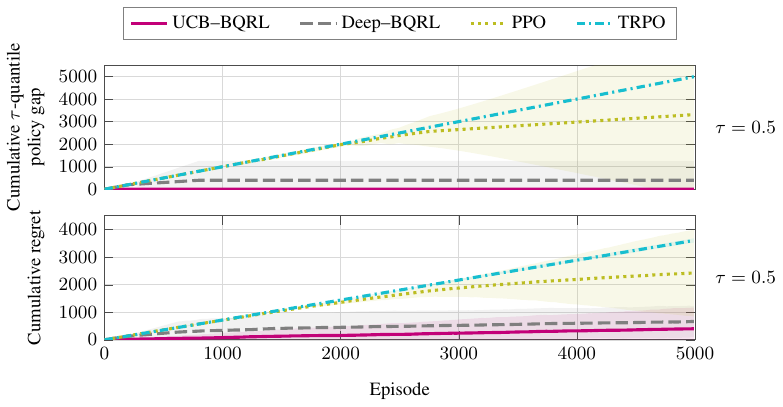}

\caption{
Learning performance on the slippery FrozenLake environment for $\tau=0.5$. The top panel reports the cumulative $\tau$-quantile policy gap, while the bottom panel reports cumulative regret under the expected-return criterion. Lower values indicate better performance.
}
\label{fig:frozenlake-slippery-regret}
\end{figure}

The binary return limits what point-quantile comparisons can reveal. For a policy with success probability $p_\pi$,
$Q_\tau(R^\pi)=0$ when $\tau\le1-p_\pi$ and $1$ otherwise. Point-quantile performance therefore changes only when the success
probability crosses $1-\tau$. At $\tau=0.1$, the cumulative point-quantile gaps are zero for all evaluated policies and do not distinguish the methods. We therefore use $\tau=0.5$ for the cumulative comparisons in Figure~\ref{fig:frozenlake-slippery-regret}. At episode $5000$, the mean cumulative point-quantile gaps are $9.67$ for \textsc{UCB--BQRL}, $400.0$ for \textsc{Deep--BQRL}, $3313.33$ for PPO, and $5000.0$ for TRPO. The \textsc{Deep--BQRL} gap shows essentially no further increase after approximately $1000$ episodes. The corresponding mean cumulative expected-return regrets are $404.28$, $666.28$, $2426.62$, and $3613.28$, respectively. Thus, \textsc{Deep--BQRL} ranks second to the model-based reference under both cumulative measures, improving on the tested policy-gradient baselines while maintaining competitive episodic reward. These results evaluate the complete method rather than isolate the effects of neural function approximation, buffered action selection, or the exploration bonus.

\section{Discussion}
\label{sec:discussion}

The results support the feasibility of learning risk-sensitive decisions from return quantiles without explicit transition estimation or return-law planning. In asset selling, \textsc{Deep--BQRL} achieves smaller mean cumulative point-quantile policy gaps than PPO and TRPO at both reported target levels, although \textsc{UCB--BQRL} remains substantially stronger. FrozenLake provides complementary evidence: \textsc{Deep--BQRL} approaches the model-based reference in episodic reward and outperforms the policy-gradient baselines in the median point-quantile comparison. Overall, these findings support \textsc{Deep--BQRL} as a practical quantile-sensitive approach, but not as a uniformly better alternative to model-based planning.

The asset-selling results also show that risk-sensitive performance and training rewards should be evaluated separately. At $\tau=0.1$, \textsc{Deep--BQRL} has a smaller mean point-quantile policy gap than PPO and TRPO but a larger expected-return regret; at $\tau=0.9$, it has the smallest mean expected-return regret. Because the point-quantile gap evaluates the greedy policy while expected-return regret includes exploration, these differences may reflect both the learned policy and data collection rather than the risk preference alone. Evaluating both expectation and point quantile for the same final policy would better separate these effects.

FrozenLake provides a different comparison because its binary return is either zero or one. A better policy therefore has a higher probability of reaching the goal, which improves expected return and cannot worsen the point or lower-buffered quantile. Thus, FrozenLake tests learning under stochastic transitions and sparse rewards, but different target quantiles need not produce different optimal policies. Moreover, once the goal-reaching probability exceeds $0.5$, the median return is one, so further improvements are not captured by the median-gap measure. Hence, a flat median-gap curve does not necessarily indicate the highest attainable success probability.

A methodological strength of \textsc{Deep--BQRL} is the direct link between the predicted return distribution and action selection. Each critic estimates fixed return quantiles, and the buffered score combines those immediately below the target level, incorporating the target quantile directly into decisions learned from sampled transitions. The method requires neither transition-model estimation nor construction of achievable return distributions during training. The controlled benchmarks also allow the learned policies to be evaluated under the true environment model rather than only through the critics' predictions. The difference between the learning criterion and evaluation measure is important. \textsc{Deep--BQRL} selects actions using $Q_\tau^\beta$, while the reported policy gap evaluates $Q_\tau$. With a nonzero buffer, these criteria may rank policies differently, and choosing actions by conditional buffered scores does not guarantee an optimal episodic point-quantile policy. The results therefore support the usefulness of the approximation empirically, rather than guarantee point-quantile optimality. The choices of $\beta$ and $K$ also require further study: $\beta$ determines which part of the return distribution is used for action selection, while $K$ controls how finely that region is represented. Sensitivity analysis could clarify their effects and whether buffering improves learning over using a single estimated quantile.

A key direction for future work is to evaluate \textsc{Deep--BQRL} in larger, higher-dimensional environments. The current experiments use small state spaces where \textsc{UCB--BQRL} remains computationally feasible, so they do not establish scalability. Experiments on Atari games and other high-dimensional discrete-control benchmarks would test whether \textsc{Deep--BQRL} retains its quantile-sensitive behavior when neural function approximation is necessary. Future studies should also consider longer horizons, larger action spaces, computational cost, and sensitivity to the buffer width, number of quantiles, and ensemble size.

\bibliography{references}
\bibliographystyle{references}

\end{document}